\documentclass{article}

\usepackage[final,nonatbib]{neurips_2019}

\usepackage[utf8]{inputenc} 
\usepackage[T1]{fontenc}    
\usepackage{hyperref}       
\usepackage{url}            
\usepackage{booktabs}       
\usepackage{amsfonts}       
\usepackage{nicefrac}       
\usepackage{microtype}      
\usepackage{cite}
\usepackage{amsmath}
\usepackage{multirow}
\usepackage{graphicx}
\usepackage{adjustbox}
\usepackage[table,xcdraw]{xcolor}
\usepackage{subfigure}
\usepackage{float}
\usepackage{cleveref}

\title{Variational Deep Survival Machines:\\
Survival Regression with Censored Outcomes
}

\author{%
  Qinxin Wang\\
  Carnegie Mellon University\\
  \texttt{qinxinw@andrew.cmu.edu} \\
  \And
  Jiayuan Huang \\
  Carnegie Mellon University \\
  \texttt{jiayuan3@andrew.cmu.edu} \\
  \AND
  Junhui Li \\
  Carnegie Mellon University \\
  \texttt{junhuil@andrew.cmu.edu} \\
  \And
  Jiaming Liu \\
  Carnegie Mellon University \\
  \texttt{jiaming2@andrew.cmu.edu} \\
}

\begin{document}

\maketitle 
\begin{abstract}
  Survival regression aims to predict the time when an event of interest will take place, typically a death or a failure. A fully parametric method \cite{DSM} is proposed to estimate the survival function as a mixture of individual parametric distributions in the presence of censoring.
  In this paper, We present a novel method to predict the survival time by better clustering the survival data and combine primitive distributions. We propose two variants of variational auto-encoder (VAE), discrete and continuous, to generate the latent variables for clustering input covariates. The model is trained end to end by jointly optimizing the VAE loss and regression loss.
  Thorough experiments on dataset SUPPORT and FLCHAIN show that our method can effectively improve the clustering result and reach competitive scores with previous methods. We demonstrate the superior result of our model prediction in the long-term. Our code is available at \url{https://github.com/qinzzz/auton-survival-785}
  

\end{abstract}

\section{Introduction}
Survival Analysis is a machine learning problem for analyzing the expected duration of time until an event occurs, typically the death of a patient in medical treatment. It has been extensively used in many medical applications. Formally, the problem can be described as a regression problem to estimate the conditional survival curve $P(T>t|X)$, which T is the event of interest and $X$ is covariate. Given $X$, the survival function predicts the probability of an event $T$ that happens after time $t$. In practice, the presence of censoring, where the outcome measurement is only partially known during the whole experiment, makes the survival analysis more difficult. For example, if the study ends before the event happens, we would not know the exact outcome for this individual's data collection. The censored observations cannot be simply discarded because they contain crucial information \cite{ref5}. \cite{watt1996survival} has also shown the importance of censored instances for deriving the accurate estimation.

Traditional statistical approaches are usually non-parametric or semi-parametric for survival regression. However, non-parametric methods will face curse of dimensional problem. Meanwhile semi-parametric methods depend on strong modelling assumptions which are usually unrealistic. Recently, some fully parametric deep learning approaches have been proposed for survival regression. For example, Deep Survival Machine\cite{DSM} estimates the survival function as a mixture of individual parametric distributions in the presence of censoring and it doesn't need strong assumptions. \cite{ref4} generalizes DSM by assuming within each latent group the proportional hazards assumption holds.

Clustering is also a valuable tool in survival analysis. Patients can usually be clustered into different latent groups and each of these groups would have an individual distribution. \cite{DBLP} presents a novel method for clustering survival data — variational
deep survival clustering (VaDeSC) that discovers groups of patients characterised by different generative mechanisms of survival outcome.

In this project, we propose two models named \textit{VDSM-cat} and \textit{VDSM-clus}, which are novel approaches combine DSM and VAE to both cluster data and predict survival time. The input data will first pass through VAE's encoder and we will get latent variables. These latent variables will be transformed as weights of DSM's mixed individual distributions. Compared to original DSM, our model achieve better results in the long term, which suggests the embedded VAE could give a better cluster outcome and is helpful for final prediction.

\textbf{Our main contributions} are as follows:

({\romannumeral1}) We propose two new methods VDSM-cat and VDSM-clus, which corresponds to categorical distribution and Gaussian Mixture distribution for the latent variable respectively. 

({\romannumeral2}) We compare our approaches with DSM on \textit{SUPPORT} and \textit{FLCHAIN} dataset and find our methods could achieve superior results in long time prediction.




\section{Related Works} 
\label{gen_inst}

Time-to-event analysis with censoring is an important problem which could be applied into many domains including bio-statistics and health informatics\cite{7822579}, actuarial sciences\cite{czado2002application} and econometrics\cite{stepanova2002survival}. It has also been increasingly popular in engineering, including applications to predict maintenance time of equipments. \cite{RePEc:eee:reensy:v:172:y:2018:i:c:p:25-35} proposed a Bayesian hierarchical model to predict the end of life (EoL) and end of discharge (EoD) of Li-ion batteries using load profiles and discharge data. \cite{ghasemi2007optimal} used survival regression model to track the degradation of the system and plan optimal maintenance strategy.

Research in survival analysis pays more attention to deep learning approaches in recent years \cite{Nagpal2022CounterfactualPW} \cite{Katzman2016DeepSA}. \cite{Lee2018DeepHitAD} proposed a fully parametric model named \textit{Deephit} to predict survival outcomes in competing risks situation. However, their architecture could only predict failure times over a discrete set of fixed size. When survival horizons become long, this approach would require large number of parameters to be learnt, which can be very impractical. Another limitation of this approach is that it can be more sensitive to events at shorter horizons and it will not model long term horizon well. \cite{lee2019temporal} propose using black box optimization to adaptively select the best model for a given event horizon.
Deep Survival Machine (DSM) \cite{DSM} is proposed as a fully-parametric approach to estimate time-to-event in the presence of censoring and competing risks. 
In DSM, the survival function is estimated as a mixture of individual parametric survival distributions, where the distributions parameters are learnt by training a Multilayer Perceptron. The strong assumptions of proportional hazards is not required. In this way, it enables the model to learn a rich distributed representation of the input covariates. 
\cite{ref4} proposed a new approach based on learning mixtures of Cox regressions to model individual survival distributions. It involves estimating hazard ratios within latent clusters followed non-parametric estimation of the baseline survival rates.

Semi-supervised clustering for survival data has been first studied in \cite{bair2004semi}. It proposed pre-selecting variables based on univariate Cox regression hazard scores and then performing k-means clustering on the subset of features to discover patient subpopulations. \cite{ahlqvist2018novel} uses Cox regression to explore differences across subgroups of diabetic patients discovered by hierarchical clustering. \cite{mouli2018deep} propose a deep clustering approach to differentiate between long- and short- term survivors based on a modified Kuiper statistic in the absence of end-of-life signals.\cite{xia2019outcome} used a multitask learning approach for the outcome-driven clustering of acute syndrome patients. In \cite{DSM} and \cite{ref4}, DSM and DCM set a mixture of survival models through representations learnt by an encoder neural network.Recently, Variational AutoEndoers (VAE) has also been involved in time-to-event analysis and clustering. \cite{ref5} introduces a varitional time-to-event prediction model, which uses Variational Survival Inference (VSI) to predict the time-to-event distribution without the need to specify a parametric form for the baseline distribution. In this way, they avoid the restrictive assumptions in classical survival analysis models. \cite{DBLP} proposed a semi-supervised approach to cluster survival data. It outperforms baseline methods through clustering the latent variables and is comparable in terms of time-to-event predictions.





\section{Our Method}
\subsection{Problem Definition}
We consider a dataset of right censored observations $\mathcal{D} = \{(x_i, \delta_i, u_i)\}^N_{i=1}$, where $x_i$ are the covariates of an individual i, $\delta_i$ is an indicator of whether an event occurred or not and $u_i$ is either the time of event or censoring as indicated by $\delta_i$. Our goal is to learn a survival function $S(t|x) = \mathbb{P}(T > t| X = x) $ from the input data. We denote the uncensored subset $(\delta=1)$ of data as $\mathcal{D_U}$ and the censored subset $(\delta=0)$ as $\mathcal{D_C}$. 

\subsection{Deep Survival Machine}
In this section, we briefly describe the model proposed by Deep Survival Machine (DSM)~\cite{DSM}. Given the above variable $x$ and $z$, the survival time $t$ can be modeled as 
\begin{equation}
   p(T|X) = \sum_Z p(T|X,Z) p(Z|X) 
\end{equation}

The model aims to maximize the probability over the entire dataset. Considering both censored and uncensored data, we model the probability by 
\begin{equation}
    p(T|X,Z) = \prod_{i=1}^{|D|} p(T=t_i|Z)^{\delta_i} p(T>t_i|Z)^{1-\delta_i}
\end{equation}
And the loss can be expressed as
\begin{equation}
    \mathcal{L}_{DSM} = \ln p(T|X,Z) = \sum_{i=1}^{|D|} \delta_i \ln p(T=t_i|Z) + \sum_{i=1}^{|D|} (1-\delta_i) \ln p(T>t_i|Z)
\end{equation}
We denote it as combining two separate loss for uncensored and censored data:
\begin{equation}
    \mathcal{L}_{DSM} = \mathcal{L}_{U} + \mathcal{L}_{C}
\end{equation}

The log likelihood of the uncensored dataset can be denoted as

\begin{equation}
    \begin{split}
    \mathcal{L}_{U} = \ln P(\mathcal{D_U};\Theta)
    &= \ln \prod_{i=1}^{|D_U|} \mathbb{P} (T=t_i |X = x_i;\Theta) \\
    &= \sum_{i=1}^{|D_U|} \ln (\sum_{k=1}^{K} \mathbb{P} (T=t_i |Z) \mathbb{P}(Z | X=x_i;w)) \\
    &= \sum_{i=1}^{|D_U|} \ln (\mathbb{E}_{Z \sim p(z|x_i,w)} \mathbb{P} (T=t_i |Z)) \\ 
    &\ge \sum_{i=1}^{|D_U|} (\mathbb{E}_{Z \sim p(z|x_i,w)} [\ln \mathbb{P} (T=t_i |Z)])
    \end{split}
\end{equation}

For the censored data, we  maximize the log likelihood by computing the probability of event happening after $t_i$.
\begin{equation}
    \begin{split}
    \mathcal{L}_{C} = \ln P(\mathcal{D_C};\Theta)
    &= \ln \prod_{i=1}^{|D_C|} \mathbb{P} (T>t_i |X = x_i;\Theta) \\
    &\ge \sum_{i=1}^{|D_C|} (\mathbb{E}_{Z \sim p(z|x_i,w)} [\ln \mathbb{P} (T>t_i |Z)])
    \end{split}
\end{equation}



\subsection{Proposed Model}
As in DSM, we choose to model the conditional distribution of $\mathbb{P}(T|X)$ as a mixture over K well-defined distributions. 
We refer to them as primitive distributions. To satisfy certain properties of survival time prediction, DSM experimented with two types of distributions, the Weibull and the Log-Normal distribution (see Table~\ref{tab:weibull}). For each primitive distribution, we need a independent set of parameters $\{\eta_k, \beta_k\}$. The final individual survival distribution $\mathbb{P}(T|X)$ is a weighted average over these K distributions.

\begin{figure*}[!htb]
    \centering
    
    \includegraphics[width=0.8\textwidth]{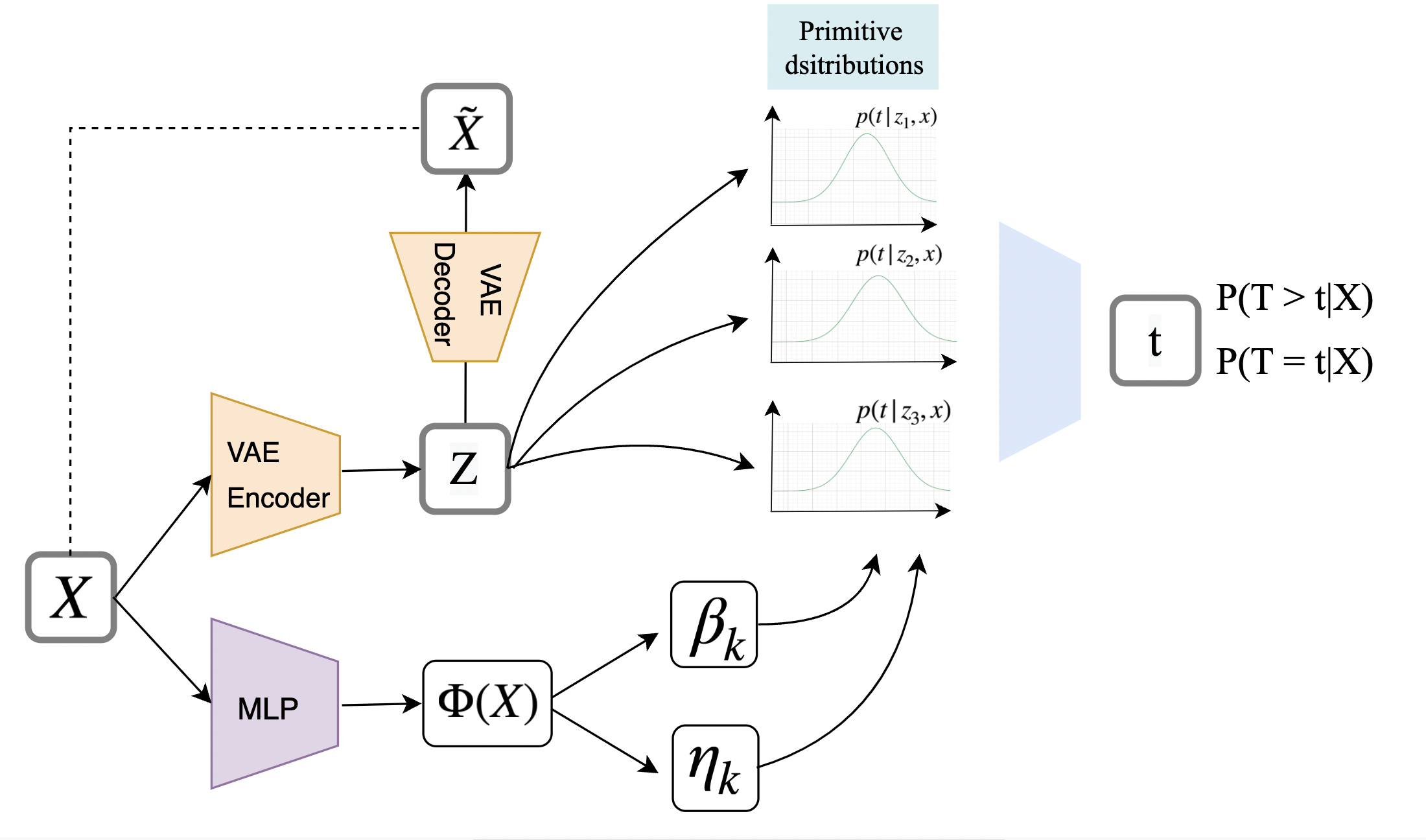}
    \caption{
    Overview of our proposed VDSM.}
    \label{fig:overview}
\end{figure*}

In DSM paper, they use a linear transform $x^T W$ to get a $z$ with $K$ classes. Though simple and intuitive, this classification method fails to learn a good weight over K categories in a unsupervised setting.
We propose a variant of DSM\cite{DSM}, Variational Deep Survival Machine (VDSM), by introducing Variational AutoEncoder\cite{Kingma2014AutoEncodingVB} into the stage of sampling $z$ from latent space given input $x$. Our goal is to learn a better cluster assignment with VAE models.
Figure \ref{fig:overview} shows an overview of the model structure. The input X is a set of features an individual patient. We use a VAE encoder to generate the latent variable Z. The Z constains of K classes, where each cluster has an independent underlying survival distribution. The final predictied distribution is $\mathbb{P}(T|X)$ is a weighted average over the K distributions.

In the VAE encoder and decoder module, we experimented with two different VAE for classification and clustering respectively. We refer to them as VDSM-cat and VDSM-clus.

{
\begin{table}[!htb]
    \large
    \caption{Choices of primitive distributions}
    \label{tab:weibull}
    \begin{adjustbox}{width=.5\textwidth,center} 
    \begin{tabular}{lcccc}
    \toprule
    & Weibull & Log-Normal \\
    \midrule
    PDF(t) & $\frac{\eta}{\beta} (\frac{t}{\beta})^{\eta-1} e^{-\frac{t}{\beta}^{eta}}$  & $\frac{1}{t\beta\sqrt{2\pi}} e^{-\frac{(\ln{t}-\eta)^2}{2\beta^2}}$ \\
    CDF(t) & $e^{-\frac{t}{\beta}^{eta}}$  & $\frac{1}{2} \text{erfc} (-\frac{\ln{t}-\eta}{\sqrt{2}\beta})$ \\
    \bottomrule
    \end{tabular}
    \end{adjustbox}
\end{table}
}

\paragraph{Variational Deep Survival Machine with categorical VAE (VDSM-cat)}
\begin{figure*}[!htb]
    \centering
    
    \includegraphics[width=0.55\textwidth]{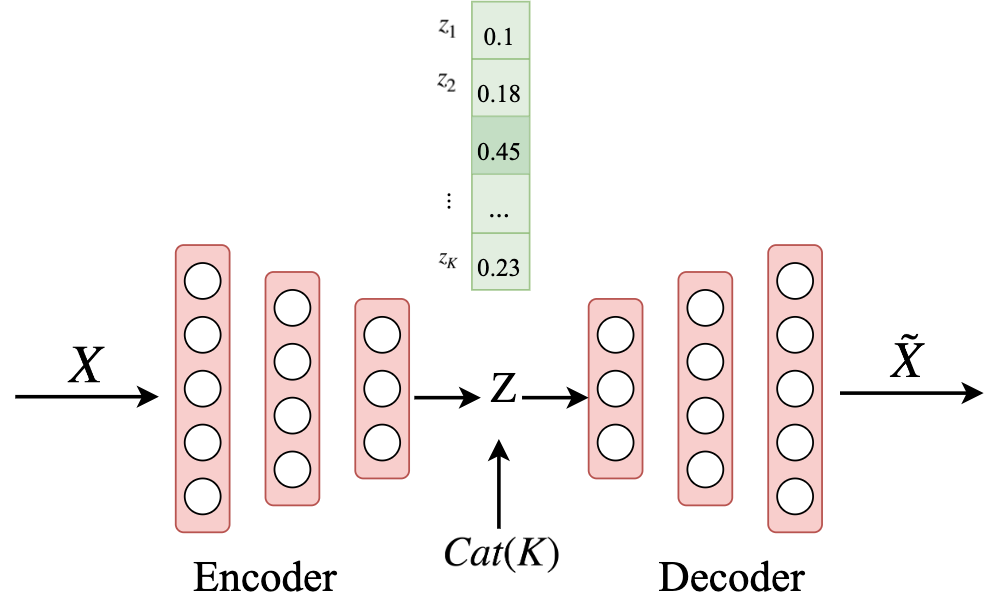}
    \caption{
    The structure of categorical VAE.}
    \label{fig:cateVAE}
\end{figure*}
We introducing Categorical Variational AutoEncoder proposed in \cite{Jang2017CategoricalRW} into the stage of sampling $z$ from latent space given input $x$, which is illustrated in Figure \ref{fig:cateVAE}.
Consider a latent space with $K$ clusters $\{ 1, ..., K \}$, $z$ is sampled from a categorical distribution $z \sim p(z) = Cat(K; \pi)$.
We know that categorical distribution is discrete and not differentiable. To make the model amenable to gradient based optimization, we use the Gumbel-Max trick\cite{Gumbel1954StatisticalTO} to generate samples which can be used in backpropagation.The Gumbel-Max trick is an efficient method to generate samples from a categorical distribution:


\begin{equation}
    z = \textrm{one\_hot}(\mathop{\arg\min}\limits_{i}[g_i+\log{\pi_i}])
\end{equation}

where $g_i$ is the samples drawn from $Gumbel(0,1)$, $\pi_i$ is the ith-class possibility of the categorical distribution. To cover the gap that argmax is not differentiable, the Gumbel-Softmax\cite{Jang2017CategoricalRW} is proposed as a continuous approximation:

\begin{equation}
    y_i = \frac{exp((\log{(\pi_i)}+g_i)/ \tau)}
    {\sum\nolimits_{j=1}^k exp((\log{(\pi_j)}+g_j)/ \tau)}  \quad\quad\quad \textrm{for }i = 1,...,k.
\end{equation}
\\
\\The loss for this categorical VAE would be the Kullback-Leibler divergence between generated distribution of $z$ with the expected categorical distribution $p(z)$, plus the reconstruction loss of $X$.
\begin{equation}
    \mathcal{L}_{VAE} = \mathbb{KL}(p(z)||q(z|x)) + \mathbb{E}_{z \sim q(z|x)} [\log p(x|z)]
\end{equation}

\paragraph{Variational Deep Survival Machine with generative clustering (VDSM-clus)} 

\begin{figure*}[!htb]
    \centering
    
    \includegraphics[width=0.6\textwidth]{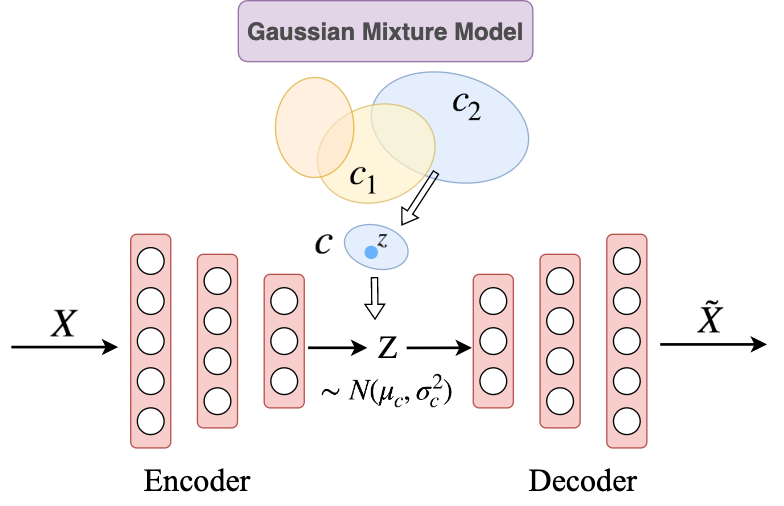}
    \caption{
    The structure of generative clustering VAE.}
    \label{fig:vade}
\end{figure*}

The assumption behind categorical VAE is that $x$ belongs to one of the $K$ categories. However, the latent variable of VAE assumes a Gaussian mixture as prior. We believe the latent cluster assumption of Gaussian Mixture Model (GMM) can be applied to extend it with a mixture of Gaussian distributions. 
Here we first describe the process of the generative model. A cluster $c$ is sampled from a categorical distribution $c \sim p(c) = Cat(\pi) $. Every cluster is a Gaussian distribution with mean $\mu_c$ and variance $\sigma_c$. Then a continuous latent embedding $z$ is sampled from the Gaussian Distribution. The input covariant $x$ is generated from a distribution conditioned on $z$. Finally, the survival time $t$ depends on $z$ and $c$, i.e., $t \sim p(t|z,c)$.
The main difference here is that $z$ conforms to a normal distribution depends on cluster $c$.

The generative process above can be defined by the following,

\begin{align}
    p(x, z, c)=p(x|z)p(z|c)p(c)
\end{align}
where
\begin{align}
    p(c) &= Cat(c, \pi)\\
    p(z|c) &= \mathcal{N}(z|\mu_c, \sigma_c^2)\\
    p(x|z) &= \mathcal{N}(x|\mu_x, \sigma_x^2)
\end{align}

\cite{Jiang2017VariationalDE} proposed an unsupervised generative clustering framework VaDE that combines VAE and GMM together, as is shown in Figure~\ref{fig:vade}. Inspired by their method, we optimized the VAE by maximizing the evidence lower bound (ELBO) using the SGVB estimator and the reparameterization trick.
We borrowed the learning objective derived in the paper, which is defined as:
\begin{align}
    \label{equ:elbo}
    \mathcal{L}_{VAE} &= \mathcal{L}_{ELBO}(x) \\
    & = E_{q(z,c|x)} [ \log \frac{p(x,z,c)}{q(z,c|x)} ] \\
    &= E_{q(z,c|x)} [\log p(x|z)] - D_{KL}(q(z,c|x)||p(z|c))
\end{align}
The first term in Equation~\ref{equ:elbo} is the reconstruction term between encoder input and decoder output. The second term is the Kullback-Leibler divergence between the Mixture-of-Gaussian prior p(z|c) to the posterior q(z,c|x). The training procedure aims to maximize the ELBO with regard to parameters. The cluster assignment is obtained by the approximation of q(c|x):
\begin{align}
    q(c|x) = p(c|z) = \frac{p(c) p(z|c)}{\sum_{c'=1}^{K} p(c') p(z|c')}
\end{align}

Finally, the combined loss of our two proposed models are:
\begin{equation}
\begin{split}
    \mathcal{L}
    &= \mathcal{L}_{DSM} + \mathcal{L}_{VAE} \\
\end{split}
\end{equation}


\section{Experimental Setup}

\subsection{Baseline}

\paragraph{Deep Survival Machine}
Deep Survival Machine is a novel approach to estimate time-to-event in the presence of censoring. By leveraging a hierarchical graphical model parameterized by
neural networks, the model learns distributional representations of the input covariates and mitigate existing challenges in survival regression. DSM estimates the conditional survival function
$\mathbb{S}(\cdot \mid X)$ as a mixture of individual parametric survival distributions without a strong assumptions of proportional hazards, and enables learning with time-varying risks. In this way, DSM allows for learning of rich distributed representations of the input covariates, helping knowledge transfer across multiple competing risks.



 
\subsection{Dataset}

\paragraph{SUPPORT} (Study to understand prognoses and preferences for outcomes and risks of
~\cite{knaus1995support}):
This Dataset is composed of 9,105 terminally ill patients on life support. The median of studied patients' survival time was 58 days. A majority with 79\% population were recorded as ‘White’, while the rest were coded as ‘Black’, ‘Hispanic’ and ‘Asian'.

\paragraph{FLCHAIN} (Assay of Serum Free Light Chain): This is a public dataset developed by \cite{Dispenzieri} aiming to study the relationship between serum free light chain and mortality. It includes covariates like age, gender, serum creatinine and presence of monoclonal gammapothy. We used a subset of this dataset, where all the individuals with missing covariates are removed. The remaining dataset consists of 6,524 individuals.

\begin{table}[htb]
\centering
\caption{Concordance-Index on \textbf{SUPPORT} dataset at different quantiles of event times}
\label{table2}
\begin{tabular}{
>{\columncolor[HTML]{FFFFFF}}c |
>{\columncolor[HTML]{FFFFFF}}c 
>{\columncolor[HTML]{FFFFFF}}c 
>{\columncolor[HTML]{FFFFFF}}c }
\toprule
\cellcolor[HTML]{FFFFFF} & \multicolumn{3}{c}{\cellcolor[HTML]{FFFFFF}Time-dependent Concordance-Index} \\ \cmidrule{2-4}
\cellcolor[HTML]{FFFFFF} & \multicolumn{3}{c}{\cellcolor[HTML]{FFFFFF}Quantiles of Event Times}         \\ \cmidrule{2-4} 
\multirow{-3}{*}{\cellcolor[HTML]{FFFFFF}Models} & 25\%                     & 50\%                     & 75\%                     \\ \cmidrule{1-4}
DSM                                              & \textbf{0.7758 ± 0.0013} & \textbf{0.7085 ± 0.0023} & 0.6560 ± 0.0032          \\ \cmidrule{1-4}
VDSM-cat                                         & 0.7580 ± 0.0025          & 0.6920 ± 0.0012          & 0.6734 ± 0.0023          \\ \cmidrule{1-4}
VDSM-clu                                         & 0.7585 ± 0.0011          & 0.6923 ± 0.0055          & \textbf{0.6736 ± 0.0022} \\ \bottomrule
\end{tabular}

\end{table}

\begin{table}[htb]
\centering
\caption{ROC-AUC on \textbf{SUPPORT} dataset at different quantiles of event times}
\label{table3}
\begin{tabular}{
>{\columncolor[HTML]{FFFFFF}}c |
>{\columncolor[HTML]{FFFFFF}}c 
>{\columncolor[HTML]{FFFFFF}}c 
>{\columncolor[HTML]{FFFFFF}}c }
\toprule
\cellcolor[HTML]{FFFFFF} & \multicolumn{3}{c}{\cellcolor[HTML]{FFFFFF}ROC-AUC} \\ \cmidrule{2-4}
\cellcolor[HTML]{FFFFFF} & \multicolumn{3}{c}{\cellcolor[HTML]{FFFFFF}Quantiles of Event Times}         \\ \cmidrule{2-4} 
\multirow{-3}{*}{\cellcolor[HTML]{FFFFFF}Models} & 25\%                     & 50\%                     & 75\%                     \\ \cmidrule{1-4}
DSM                                              & \textbf{0.7841 ± 0.0017} & \textbf{0.7298 ± 0.0033} & 0.7097 ± 0.0038          \\ \cmidrule{1-4}
VDSM-cat                                         & 0.7672 ± 0.0025          & 0.7127 ± 0.0054          & 0.7212 ± 0.0004          \\ \cmidrule{1-4}
VDSM-clu                                         & 0.7677 ± 0.0011          & 0.7130 ± 0.0027          & \textbf{0.7215 ± 0.0015} \\ \bottomrule
\end{tabular}
\end{table}

\subsection{Experiment Metrics}
We evaluate the baseline model DSM, the proposed model VDSM-cat and VDSM-clus with two popular metrics in survival analysis literature. Similar to DSM settings \cite{DSM}, we evaluate these two metrics with different truncation event horizon quantiles of $25\%$, $50\%$ and $75\%$ to measure how good the model can be at showing the risks at different scale of time.

\paragraph{Concordance index $C^{td}$} It is used to evaluate the proportion of all comparable pairs in which the predictions and outcomes are concordant. More specifically, Concordance intuitively means that two samples were ordered correctly by the model. If the one with a higher estimated risk score has a shorter actual survival time, we can say the two samples are concordant.

\paragraph{ROC-AUC} In survival analysis, the area under receiver operating characteristic (ROC) curve is used by defining sensitivity and specificity as time-dependent measures. All individuals that experienced an event prior to or at time $t$ are Cumulative cases and dynamic controls are those individuals experienced an event after the time.

\subsection{Hyper-parameter Searching}
To achieve the best performance, we use grid search to find the best hyper-parameter settings for the DSM model. As reported in the DSM paper \cite{DSM}, we evaluate the model with the following parameter settings: the number of clusters is chosen from $\{4,6,8\}$, the discounting factor is chosen from $\{0.5,0.75,1\}$, and the learning rate of optimizer Adam is chosen from $\{1e-3, 1e-4\}$. The best performance is achieved with $4$ clusters, $0.5$ as the discounting factor and $1e-4$ as the learning rate.

For our proposed VDSM-cat and VDSM-clus, we research the hyper-parameter settings. VAEs are added to the DSM to get better clustering representation and therefore it may behave differently with different number of clusters. 
Our experiment shows that the VDSM-clus outperforms the VDSM-cat with every cluster number in $\{4,6,8\}$ according to the Concordance index and ROC-AUC. For SUPPORT dataset, $10$ clusters gives the best result and for FLCHAIN dataset, $6$ clusters shows the best performance.

\begin{table}[htb]
\centering
\caption{Concordance-Index on \textbf{FLCHAIN} dataset at different quantiles of event times}
\label{table4}
\begin{tabular}{
>{\columncolor[HTML]{FFFFFF}}c |
>{\columncolor[HTML]{FFFFFF}}c 
>{\columncolor[HTML]{FFFFFF}}c 
>{\columncolor[HTML]{FFFFFF}}c }
\toprule
\cellcolor[HTML]{FFFFFF} & \multicolumn{3}{c}{\cellcolor[HTML]{FFFFFF}Time-dependent Concordance-Index} \\ \cmidrule{2-4}
\cellcolor[HTML]{FFFFFF} & \multicolumn{3}{c}{\cellcolor[HTML]{FFFFFF}Quantiles of Event Times}         \\ \cmidrule{2-4} 
\multirow{-3}{*}{\cellcolor[HTML]{FFFFFF}Models} & 25\%                     & 50\%                     & 75\%                     \\ \cmidrule{1-4}
DSM                                              & \textbf{0.6033 ± 0.0033} & \textbf{0.6641 ± 0.0039} & 0.6170 ± 0.0032          \\ \cmidrule{1-4}
VDSM-cat                                         & 0.5848 ± 0.0015          & 0.6229 ± 0.0016          & 0.6125 ± 0.0010          \\ \cmidrule{1-4}
VDSM-clu                                         & 0.5635 ± 0.0006          & 0.6449 ± 0.0049          & \textbf{0.6219 ± 0.0050} \\ \bottomrule
\end{tabular}

\end{table}

\begin{table}[htb]
\centering
\caption{ROC-AUC on \textbf{FLCHAIN} dataset at different quantiles of event times}
\label{table5}
\begin{tabular}{
>{\columncolor[HTML]{FFFFFF}}c |
>{\columncolor[HTML]{FFFFFF}}c 
>{\columncolor[HTML]{FFFFFF}}c 
>{\columncolor[HTML]{FFFFFF}}c }
\toprule
\cellcolor[HTML]{FFFFFF} & \multicolumn{3}{c}{\cellcolor[HTML]{FFFFFF}ROC-AUC} \\ \cmidrule{2-4}
\cellcolor[HTML]{FFFFFF} & \multicolumn{3}{c}{\cellcolor[HTML]{FFFFFF}Quantiles of Event Times}         \\ \cmidrule{2-4} 
\multirow{-3}{*}{\cellcolor[HTML]{FFFFFF}Models} & 25\%                     & 50\%                     & 75\%                     \\ \cmidrule{1-4}
DSM                                              & \textbf{0.6057 ± 0.0020} & \textbf{0.6689 ± 0.0007} & 0.6201 ± 0.0014          \\ \cmidrule{1-4}
VDSM-cat                                         & 0.5874 ± 0.0052          & 0.6260 ± 0.0048          & 0.6215 ± 0.0019          \\ \cmidrule{1-4}
VDSM-clu                                         & 0.5656 ± 0.0003          & 0.6502 ± 0.0051          & \textbf{0.6430 ± 0.0026} \\ \bottomrule
\end{tabular}
\end{table}

\section{Results and Discussion}
\subsection{Experiment Results}
We run Deep Survival Machine (DSM) \cite{DSM} and our proposed models on SUPPORT and FLCHAIN dataset. 
Our code for DSM is adapted from the open-source framework Auton-Survival~\cite{nagpal2022auton}.
The results are shown in Table~\ref{table2}, \ref{table3}, \ref{table4}, and \ref{table5}.

\subsection{Clustering Results}
We run multiple experiments on SUPPORT and FLCHAIN dataset to collect results. First, we generate the latent variable after training the model. Since the latent variable represents a distribution over latent clusters, we choose the category with the largest probability as the label and draw a plot to visualize the clustering result with the t-SNE algorithm.

\begin{figure*}[ht]
\centering
\subfigure[DSM]{\includegraphics[height=4.5cm,width=4.5cm]{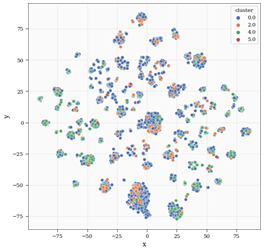}}
\subfigure[VDSM-cat]{\includegraphics[height=4.5cm,width=4.5cm]{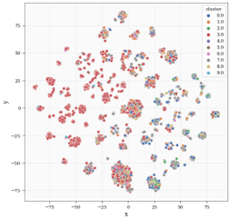}}
\subfigure[VDSM-clus]{\includegraphics[height=4.5cm,width=4.5cm]{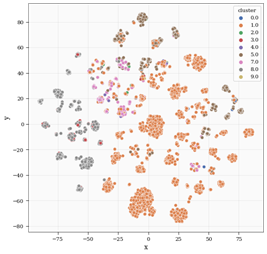}}
\caption{ Clustering Result on SUPPORT dataset }
\label{clustering_support}
\end{figure*}

\begin{figure*}[ht]
\centering
\subfigure[DSM]{\includegraphics[height=4.5cm,width=4.5cm]{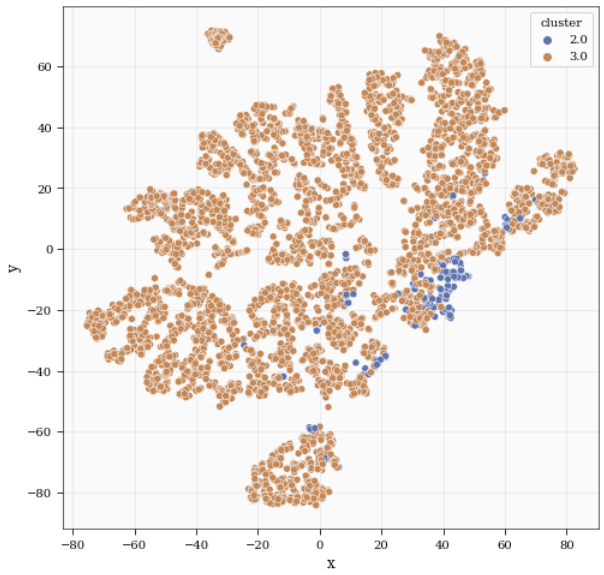}}
\subfigure[VDSM-cat]{\includegraphics[height=4.5cm,width=4.5cm]{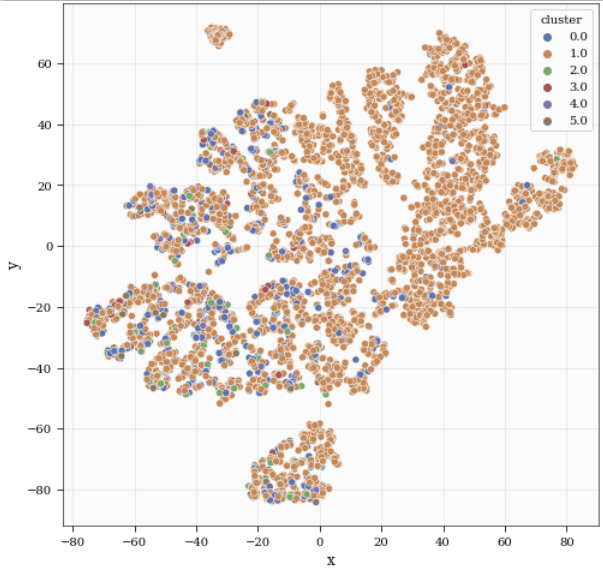}}
\subfigure[VDSM-clus]{\includegraphics[height=4.5cm,width=4.5cm]{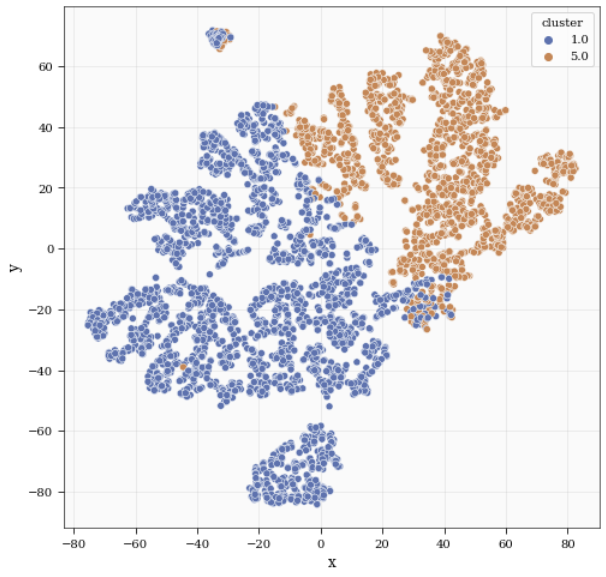}}
\caption{ Clustering Result on FLCHAIN dataset }
\label{clustering_flchain}
\end{figure*}

Figure \ref{clustering_support} and Figure \ref{clustering_flchain} show the clustering result on SUPPORT and FLCHAIN dataset, respectively. 
Compared to DSM, VDSM-clus has a better clustering result, which is not surprising because the reconstruction loss forces the VAE to learn the latent embedding better.

It should be noted that VDSM-cat does not show better clustering result. In categorical VAE, the latent variable consists of latent dimension $N$ and category dimension $K$. In our implementation, we set $K$ as the number of latent clusters in DSM, and regard $N$ as a multi-nominal distribution of with $N$ independent trails. 
To be specific, the probability of selecting category c is $ \frac{P(c |x, n=1,2,...,N)}{ \sum_{c'} P(c'|x, n=1,2,...,N)} $. Since we assume the N latent dimensions are independent, we have $ P(c|x) = P(c|x, n=1,2,...,N) = \prod_i^N P(c|x,n=i) $.
This may not be the best way to handle the latent dimension in categorical VAE and is probably the reason why the clustering result of VDSM-cat is not as expected.

\subsection{Better Prediction at Long-time Scale}
We compare the results between our models and DSM based on the Concordance Index. The Concordance Index is a commonly used metric in survival analysis, and it evaluate the probability of the pair risk of different patient, at the time one event has already occurred to one of the patients, which means, ideally, that patient should be assigned with a higher risk. The bigger this probability is, the more consistent the prediction is with the truth.

It shows the potential of VAE to capture the long-time-dependent features with the latent embedding of inputs. Another possible explanation is that the latent representation clustering is more important to survival prediction in a long time scale. We are still working on this to prove our assumptions. The ROC-AUC also shows similar results.

\subsection{Effect of Clustering on Time-to-event Prediction}
We also found that different quality of clustering does not affect the time-to-event prediction much. According Figure \ref{clustering_support} and Figure \ref{clustering_flchain}, on both SUPPORT and FLCHAIN datasets the VDSM-clus has a better result than the VDSM-cat. However, experiment results show that there is no much difference between these two models.

Manduchi et al.\cite{DBLP} also show similar results by comparing different clustering method. Their proposed model is better at clustering but only stay at competitive with other state-of-the-art survival models. Perhaps the clustering is not the best way to represent the input data.



\section{Conclusion \& Future Work}

We proposed to incorporate  Variational AutoEncoder to the Deep Survival Machines, which is a semi-supervised deep probabilistic model for clustering survival data. By experimentation, we demonstrated that the VAE model leads to a better result in clustering patients to different latent groups than the naive approach in DSM architecture. Furthermore, the grouping information allows us to get more accurate survival time prediction for patients comparing with state-of-art models. Interesting future work directions include trying other variational clustering methods for survival data, e.g. Random Forest, a Multiple Task Variational Autoencoder to better study the latent relationship between features. Furthermore, the model can be modified to be able to predict survival time based on image input. 

\section{Division of Work}
\textbf{Jiaming Liu} -  Jiaming co-researched the survival models, collected references about VAE and other deep learning models for survival regression and took part in the design of VAE-cat and VAE-clus. He pre-processed MNIST to do some early experiments and implemented several visualization tools to help analyze model results. He also wrote Introduction and Related Works part in this report.

\textbf{Jiayuan Huang} - Jiayuan co-researched the survival models used in survival analysis, and designed the method of incorporating categorical into DSM. He implemented the proposed model VDSM-cat and conducted experiments on two datasets to collect results. He also wrote the Result and Discussion part in this report.

\textbf{Junhui Li} - Junhui co-researched the survival models used in survival used in survival analysis. She transferred existing VAE methods to Pytorch framework for experiment.She also researched into potential improvement methods and wrote the Dataset, Conclusion and Future Work part in this report. \\

\textbf{Qinxin Wang} - Qinxin co-researched the survival models (DSM and DCM), and worked on the mathematical formulation of our new methods. She surveyed multiple VAEs for clustering, implemented the proposed model VDSM-clus, and conducted experiments. She wrote the Method part and also drew model diagrams in this report. \\

\bibliography{refs}{}

\begin{thebibliography}{10}

\bibitem{ahlqvist2018novel}
Emma Ahlqvist, Petter Storm, Annemari K{\"a}r{\"a}j{\"a}m{\"a}ki, Mats
  Martinell, Mozhgan Dorkhan, Annelie Carlsson, Petter Vikman, Rashmi~B Prasad,
  Dina~Mansour Aly, Peter Almgren, et~al.
\newblock Novel subgroups of adult-onset diabetes and their association with
  outcomes: a data-driven cluster analysis of six variables.
\newblock {\em The lancet Diabetes \& endocrinology}, 6(5):361--369, 2018.

\bibitem{bair2004semi}
Eric Bair, Robert Tibshirani, and Todd Golub.
\newblock Semi-supervised methods to predict patient survival from gene
  expression data.
\newblock {\em PLoS biology}, 2(4):e108, 2004.

\bibitem{czado2002application}
Claudia Czado and Florian Rudolph.
\newblock Application of survival analysis methods to long-term care insurance.
\newblock {\em Insurance: Mathematics and Economics}, 31(3):395--413, 2002.

\bibitem{Dispenzieri}
Angela Dispenzieri, {Jerry A.} Katzmann, {Robert A.} Kyle, {Dirk R.} Larson,
  {Terry M.} Therneau, {Colin L.} Colby, {Raynell J.} Clark, {Graham P.} Mead,
  Shaji Kumar, {L. Joseph} Melton, and {S. Vincent} Rajkumar.
\newblock Use of nonclonal serum immunoglobulin free light chains to predict
  overall survival in the general population.
\newblock {\em Mayo Clinic Proceedings}, 87(6):517--523, June 2012.
\newblock Funding Information: Grant Support: This work was supported in part
  by grants CA62242 (A.D, R.A.K. S.V.R.), CA107476 (A.D., S.V.R., J.A.K.,
  R.A.K.), and CA91561 (A.D) from the National Cancer Institute , The JABBS
  Foundation , and The Predolin Foundation . Binding Site provided the serum
  immunoglobulin free light chain reagent. This study was supported in part by
  National Institutes of Health grant R01 AG034676 and the Rochester
  Epidemiology Project (grant number R01-AG034676 ; Principal Investigator:
  Walter A. Rocca, MD).

\bibitem{ghasemi2007optimal}
Alireza Ghasemi, S~Yacout, and MS~Ouali.
\newblock Optimal condition based maintenance with imperfect information and
  the proportional hazards model.
\newblock {\em International journal of production research}, 45(4):989--1012,
  2007.

\bibitem{Gumbel1954StatisticalTO}
Emil~Julius Gumbel.
\newblock Statistical theory of extreme values and some practical applications
  : A series of lectures.
\newblock 1954.

\bibitem{Jang2017CategoricalRW}
Eric Jang, Shixiang~Shane Gu, and Ben Poole.
\newblock Categorical reparameterization with gumbel-softmax.
\newblock {\em ArXiv}, abs/1611.01144, 2017.

\bibitem{Jiang2017VariationalDE}
Zhuxi Jiang, Yin Zheng, Huachun Tan, Bangsheng Tang, and Hanning Zhou.
\newblock Variational deep embedding: An unsupervised and generative approach
  to clustering.
\newblock In {\em IJCAI}, 2017.

\bibitem{Katzman2016DeepSA}
Jared Katzman, Uri Shaham, Alexander Cloninger, Jonathan Bates, Tingting Jiang,
  and Yuval Kluger.
\newblock Deep survival: A deep cox proportional hazards network.
\newblock {\em ArXiv}, abs/1606.00931, 2016.

\bibitem{Kingma2014AutoEncodingVB}
Diederik~P. Kingma and Max Welling.
\newblock Auto-encoding variational bayes.
\newblock {\em CoRR}, abs/1312.6114, 2014.

\bibitem{knaus1995support}
William~A Knaus, Frank~E Harrell, Joanne Lynn, Lee Goldman, Russell~S Phillips,
  Alfred~F Connors, Neal~V Dawson, William~J Fulkerson, Robert~M Califf, Norman
  Desbiens, et~al.
\newblock The support prognostic model: Objective estimates of survival for
  seriously ill hospitalized adults.
\newblock {\em Annals of internal medicine}, 122(3):191--203, 1995.

\bibitem{lee2019temporal}
Changhee Lee, William Zame, Ahmed Alaa, and Mihaela Schaar.
\newblock Temporal quilting for survival analysis.
\newblock In {\em The 22nd international conference on artificial intelligence
  and statistics}, pages 596--605. PMLR, 2019.

\bibitem{Lee2018DeepHitAD}
Changhee Lee, William~R. Zame, Jinsung Yoon, and Mihaela van~der Schaar.
\newblock Deephit: A deep learning approach to survival analysis with competing
  risks.
\newblock In {\em AAAI}, 2018.

\bibitem{DBLP}
Laura Manduchi, Ricards Marcinkevics, Michela~Carlotta Massi, Verena Gotta,
  Timothy M{\"{u}}ller, Flavio Vasella, Marian~C. Neidert, Marc Pfister, and
  Julia~E. Vogt.
\newblock A deep variational approach to clustering survival data.
\newblock {\em CoRR}, abs/2106.05763, 2021.

\bibitem{RePEc:eee:reensy:v:172:y:2018:i:c:p:25-35}
Madhav Mishra, Jesper Martinsson, Matti Rantatalo, and Kai Goebel.
\newblock {Bayesian hierarchical model-based prognostics for lithium-ion
  batteries}.
\newblock {\em Reliability Engineering and System Safety}, 172(C):25--35, 2018.

\bibitem{mouli2018deep}
S~Chandra Mouli, Bruno Ribeiro, and Jennifer Neville.
\newblock A deep learning approach for survival clustering without end-of-life
  signals.
\newblock 2018.

\bibitem{Nagpal2022CounterfactualPW}
Chirag Nagpal, Mononito Goswami, Keith~A. Dufendach, and Artur~W. Dubrawski.
\newblock Counterfactual phenotyping with censored time-to-events.
\newblock {\em ArXiv}, abs/2202.11089, 2022.

\bibitem{DSM}
Chirag Nagpal, Xinyu Li, and Artur Dubrawski.
\newblock Deep survival machines: Fully parametric survival regression and
  representation learning for censored data with competing risks.
\newblock {\em IEEE Journal of Biomedical and Health Informatics},
  25(8):3163--3175, 2021.

\bibitem{nagpal2022auton}
Chirag Nagpal, Willa Potosnak, and Artur Dubrawski.
\newblock auton-survival: an open-source package for regression, counterfactual
  estimation, evaluation and phenotyping with censored time-to-event data.
\newblock {\em arXiv preprint arXiv:2204.07276}, 2022.

\bibitem{ref4}
Chirag Nagpal, Steve Yadlowsky, Negar Rostamzadeh, and Katherine Heller.
\newblock Deep cox mixtures for survival regression.
\newblock In {\em Machine Learning for Healthcare Conference}, pages 674--708.
  PMLR, 2021.

\bibitem{stepanova2002survival}
Maria Stepanova and Lyn Thomas.
\newblock Survival analysis methods for personal loan data.
\newblock {\em Operations Research}, 50(2):277--289, 2002.

\bibitem{watt1996survival}
DC~Watt, TC~Aitchison, RM~Mackie, and JM~Sirel.
\newblock Survival analysis: the importance of censored observations.
\newblock {\em Melanoma research}, 6(5):379--385, 1996.

\bibitem{xia2019outcome}
Eryu Xia, Xin Du, Jing Mei, Wen Sun, Suijun Tong, Zhiqing Kang, Jian Sheng,
  Jian Li, Changsheng Ma, Jianzeng Dong, et~al.
\newblock Outcome-driven clustering of acute coronary syndrome patients using
  multi-task neural network with attention.
\newblock In {\em MedInfo}, pages 457--461, 2019.

\bibitem{ref5}
Zidi Xiu, Chenyang Tao, and Ricardo Henao.
\newblock Variational learning of individual survival distributions.
\newblock In {\em Proceedings of the ACM Conference on Health, Inference, and
  Learning}, pages 10--18, 2020.

\bibitem{7822579}
Xinliang Zhu, Jiawen Yao, and Junzhou Huang.
\newblock Deep convolutional neural network for survival analysis with
  pathological images.
\newblock In {\em 2016 IEEE International Conference on Bioinformatics and
  Biomedicine (BIBM)}, pages 544--547, 2016.

\end{thebibliography}
\bibliographystyle{plain}

\end{document}